%% file: main_camera.tex
\documentclass[10pt,twocolumn,letterpaper]{article}

\usepackage{iccv}
\usepackage{times}
\usepackage{epsfig,paralist}
\graphicspath{{figs}}
\usepackage{graphicx}
\usepackage{amsmath}
\usepackage{amssymb}
\usepackage{bm}
\usepackage{algorithm}
\usepackage{algpseudocode}
\usepackage{subcaption}

\usepackage{cite}

\newtheorem{theorem}{Theorem}
\newtheorem{definition}{Definition}
\newtheorem{lemma}{Lemma}

\usepackage{dsfont}
\input{defs}

\usepackage[breaklinks=true,bookmarks=false]{hyperref}

\iccvfinalcopy % *** Uncomment this line for the final submission

\def\iccvPaperID{3772} % *** Enter the ICCV Paper ID here

\ificcvfinal\fi
\begin{document}

%%%%%%%%% TITLE
\title{Resource Constrained Neural Network Architecture Search: \\Will a Submodularity Assumption Help?}

\author{Yunyang Xiong \ \ \ \ Ronak Mehta \ \ \ \ Vikas Singh\\
University of Wisconsin Madison\\
{\tt\small yxiong43@wisc.edu\ \ \ \ ronakrm@cs.wisc.edu\ \ \ \ vsingh@biostat.wisc.edu}
}

\maketitle

\begin{abstract}
	\input{abstract.tex}
\end{abstract}
\input{intro.tex}
\section{Preliminaries}
% architecture search basic problem setup
\input{problem.tex}
\input{submod.tex}
\input{arch.tex}
\input{speedup.tex}
\input{reduce.tex}
\input{exps.tex}
\input{conclusion.tex}
\newpage
{\small
	\bibliographystyle{ieee_fullname}
	\bibliography{main}
}

\end{document}

%% file: defs.tex
\def\cA{\mathcal{A}}
\def\cB{\mathcal{B}}
\def\cV{\mathcal{V}}
\def\cS{\mathcal{S}}
\def\cN{\mathcal{N}}
\def\cL{\mathcal{L}}
\def\bR{\mathbb{R}}

\def\dI{\mathds{1}}
\def\bx{\textbf{x}}

%% file: abstract.tex
The design of neural network architectures 
is frequently either based on human
expertise using trial/error and empirical feedback
or tackled via large scale reinforcement learning strategies performed 
over distinct discrete architecture choices.
In the latter case, the optimization is often
non-differentiable and also not very amenable to derivative-free
optimization methods. Most methods in use today require sizable computational resources.
And if we want networks that additionally satisfy resource constraints, the above challenges
are exacerbated because the search must now balance accuracy with certain
budget constraints on resources.
We formulate this problem as the optimization of a set function -- we find that the 
empirical behavior of this set function often (but not always) satisfies
marginal gain and monotonicity principles -- properties
central to the idea of submodularity. Based on this observation,
we adapt algorithms within discrete optimization
to obtain heuristic schemes for neural network architecture search, where we have
resource constraints on the architecture.
This simple scheme
when applied on CIFAR-100 and ImageNet,
identifies resource-constrained architectures 
with quantifiably better performance
than current state-of-the-art models designed for mobile devices.
Specifically, we find high-performing architectures with fewer parameters and
computations by a search method that is much faster.

%-cnns are big
%
%-resources are limited
%
%-automated arch search a new thing, practical here
%
%-using a key observation
%
%-simple results from submodular optimzation
%
%-arch scheme that efficiently finds constant factor close optimal
%
%-applied to cifar and imagenet
%
%-best results based on mobile device constraints compared to current sota models.

%% file: intro.tex
\section{Introduction}
The design of state-of-the-art neural network architectures for a given learning task typically involves
extensive effort: human expertise as well as significant compute power. 
%from a large number of researchers.
It is well accepted that the trial-and-error process is tedious,
often requiring 
iterative adjustment of models based on empirical feedback -- 
until one discovers the ``best'' structure, often based on overall accuracy. In other cases
it may also be a function of the model's memory footprint or speed at test time. 
%by hand until discovering the best structure.
This architecture search process becomes more challenging when we seek {\em resource-constrained} networks to eventually deploy on small form factor devices: accuracy and
resource-efficiency need to be carefully balanced.
Further, each type of mobile device has its own hardware idiosyncrasies and may require
different architectures for the best accuracy-efficiency trade-off. Motivated
by these considerations, researchers are devoting
effort into the development of algorithms 
%are paying more attention to developing algorithms to
that automate the process of architecture search and design.
Many of the models that have been identified via a judicious use of such architecture search schemes (often together
with human expertise)
currently provide excellent performance in classification \cite{baker2016designing, zoph2017learning,real2018regularized}
and object detection\cite{real2018regularized}.  

%Although the automatically identified architectures has achieved remarkable performance,
The superior performance of the architectures identified via the above process notwithstanding, it is well known that
those search algorithms are time consuming and compute-intensive. 
%the developed search algorithms are computation-intensive and time-consuming.
For reference,
%\ref{??} needed 3150 GPU days to train an evolutionary-style algorithm to to find a state-of-the-art architecture. E
even for a
smaller dataset such as CIFAR-10, \cite{zoph2017learning} requires 3150 GPU days for a reinforcement learning (RL) model.
A number of approaches have been proposed to speed up architecture search algorithms.
Some of the strategies include adding a specific structure to the search space to reduce search time \cite{liu2018progressive,tan2018mnasnet},
sharing weights across various architectures \cite{cai2018efficient,pham2018efficient}, and imposing weight or performance prediction constraints for each distinct architecture \cite{liu2017hierarchical,brock2017smash,baker2016designing}. 
These ideas all help in various specific cases, but the inherent issue of a large search space and the associated difficulties of scalability still exists.

Notice that one reason why many search methods based on RL, evolutionary schemes, MCTS \cite{negrinho2017deeparchitect}, SMBO \cite{liu2018progressive} or Bayesian optimization\cite{kandasamy2018neural} are
compute intensive in general is because architecture search is often set up as a black-box optimization problem over a large discrete domain,
thus leading to a large number of architecture evaluations during search.
Further, many architecture search methods \cite{zoph2017learning,liu2018progressive,dong2018dpp} do not directly take into account certain resource bounds (e.g., \# of FLOPs) although
the search space can be pre-processed to filter out those regions of the search space. 
As a result, few methods have been widely used for identifying deployment-ready architectures for 
%making the models found hard to deploy on
mobile/embedded devices. When the resource of interest is memory, one 
choice is to first train a network and then squeeze or compress it for a target deployment device \cite{yang2018netadapt,gordon2018morphnet,ashok2017n2n,he2018amc, wang2019haq}. 

{\bf Key idea.} Here, we take a slightly different line of attack for this problem which is based loosely on ideas that were widely used in computer
vision in the early/mid 2000s \cite{Kolmogorov:2002:EFM:645317.649315}.
First, we move away from black box optimization for architecture search, similar to
strategies adopted in other recent works \cite{negrinho2017deeparchitect,liu2018progressive,kandasamy2018neural,shin*2018differentiable,howard2019searching}.
Instead, we view the architecture as being composed of primitive basic blocks. The ``skeleton'' or connectivity between
the blocks is assumed fixed whereas the actual functionality provided by each block is a decision variable -- this is precisely what our search will be performed on.
%The universe of choices enumerating the type of functionality the blocks may seek to utilize is assumed to be given.
%This is based on 
%operations or modules that have been reported in the literature to be resource efficient.
The goal then is to identify the assignment of blocks so that the overall architecture satisfies two
simple properties: \begin{inparaenum}[\bfseries (a)] \item it satisfies the user provided resource bounds and \item  is accurate for the user's task of interest. \end{inparaenum}

%{\bf Intuition.} 
While we will make the statement more formal shortly, it is easy to see that the above problem can be easily viewed as a set function.
Each empty block can be assigned to a specific type of functional module. Once all blocks have been assigned to
some functional module, we have a ``proposal'' network architecture whose accuracy can be evaluated --- either by training to completion or stopping early \cite{li2017hyperband,zhang2016understanding}. A different assignment
simply yields a different architecture with a different performance profile.
In this way, the set function (where accuracy can be thought of as function evaluation) can be queried/sampled. We find that empirically, when we evaluate the behavior of this set function, it often exhibits
nice {\em marginal gain} or {\em diminishing returns properties}.
% -- for instance, 
%the benefit in terms of accuracy is smaller when adding a block to a complex network compared to when adding the same block to simpler network.
Further, the performance (accuracy) typically {\em improves} or stays nearly the same when
{\em adding} functional modules to a currently empty block (akin to
adding an element to a set).
These properties are central to {\bf submodularity}, a key concept in
many classical methods in computer vision. 
Mathematically, of course, our set function is {\bf not} submodular.
%: indeed, certain functional blocks, when combined with
%the non-convex objective function, may hurt performance, violating monotonicity. 
However, our empirical study suggests that
the set function generally behaves well. Therefore, similar to heuristic application of convex optimization techniques to nonconvex problems,
%how techniques developed for convex optimization were heuristically used
%for optimizing highly non-convex objectives common in deep neural networks,
we utilize submodular optimization algorithms for architecture search algorithm design.

{\bf Main contributions and results.} We adapt a simple greedy algorithm with performance guarantees for submodular optimization -- in this case,
we obtain a heuristic that 
%the greedy algorithm in submodular optimization
is used to optimize architecture search with respect to its validation set performance.
%In submodularity optimization literature \ref{??}, the greedy algorithm has been shown to achieve near-optimal solutions.
This design choice actually enables achieving very favorable performance relative to state-of-the-art approaches using {\bf orders of magnitude} less computation resources,
while concurrently being competitive with other recent efficient search methods, such as ENAS,   ProxylessNAS, DARTS and FBNet \cite{liu2018darts,pham2018efficient,cai2018proxylessnas,wu2019fbnet}. 
This greedy search is also far simpler to implement than many existing search methods: no controllers\cite{baker2016designing,zoph2016neural, zoph2017learning,pham2018efficient}, hypernetworks\cite{brock2017smash}, or performance predictors \cite{liu2018progressive} are required.
It can be easily extended to a number of different resource bounded architecture search applications. 
%When we transfer the identified model to ImageNet or the model is searched on ImageNet directly, our model is still comparable to MobileNetV2.

Our {\bf contributions} are:
\begin{inparaenum}[\bfseries (1)]
\item We formulate Resource Constrained Architecture Search (RCAS) as a set function optimization and design a heuristic based on ideas known to be effective for constrained submodular optimization.
  %\item We introduce a novel algorithm for efficient architecture search and it offers an architecture solution path based on the given budget.
\item We describe schemes by which the algorithm can easily satisfy constraints imposed due to specific deployment platforms ($\#$ of FLOPs, power/energy). 
\item On the results side, we achieve remarkable architecture search efficiency. On CIFAR-100, our algorithm takes only {\bf two days on 2 GPUs} to obtain an architecture with similar complexity as MobileNetV2.
  On ImageNet, our algorithm runs in 8 days on 2 GPUs and identifies an architecture similar in complexity and performance to MobileNetV2.
\item We show that the architectures learned by RCAS on CIFAR-100 can be directly transferred to ImageNet with good performance.
\end{inparaenum}

%% file: problem.tex
\subsection{Architecture Search}
We aim to include computational constraints in the design of mobile
Convolutional Neural Networks (MCNNs). Consider a finite computational budget available for a specific prediction task, $B$. The problem of designing efficient MCNNs can be viewed as seeking the most accurate CNN model that fits within said budget $B$:
\begin{align}\label{eq:bopt}
	\underset{cnn}\max \quad f(cnn) \quad \text{subject to} \quad Cost(cnn) \leq B
\end{align}
where $f$ denotes a score function, typically the validation accuracy on a held out set of samples $X_{valid}:=(\bx_i, y_i)_{i=1}^m$, i.e., $f(cnn) = \frac{1}{m} \sum_{i=1}^m \dI_{[cnn_w(\bx_i)=y_i]}$. The parameters $w$ are learned with a training set $X_{train}$, often with a surrogate cross-entropy loss and stochastic gradient descent.

{\bf Resource constraints.} Model cost is typically measured \cite{han2015learning} in two ways, analogous to algorithmic space and time complexities: first by the number of parameters,
and second by the number of computations, multiply-adds (MAdds) or floating point operations per second (FLOPs).
With this in mind, we can more concretely define the budget constraint. For a budget $B$, assume there is a corresponding maximum number of MAdds $B_m$ and number of parameters $B_p$:
\begin{align}\label{eq:fopt}
	\underset{cnn}{\max} \quad &f(cnn) \\
	\text{s.t.} \quad &MAdds(cnn) \leq B_m, \quad Param(cnn) \leq B_p \notag
\end{align}
where \textit{MAdds}$(cnn)$ denotes the number of multiply-adds and \textit{Param}$(cnn)$ denotes the number of parameters of the model $cnn$. Typical hardware constraints are given in these formats, either through physical memory specifications or processor speeds and cache limits.

Modern CNNs are built by constructing a sequence of various types of basic blocks. In network design, one may have a variety of options in types of basic blocks, block order, and number of blocks. Examples include ResNet blocks, 3x3 convolutional layers, batchnorm layers, etc. \cite{he2016deep}. Given a large set of various basic blocks, we would like to find a subset $\cS$ that leads to a well performing, low cost CNN. Blocks may have different associated costs, both in MAdds and in number of parameters. 

% Therefore the neural architecture search optimization problem can be reformulated as 
%\begin{align}\label{eq:arch_sub_opt}
%\underset{{\cS \subseteq \cV}}{\text{Maximize}} &\quad F(\cS) \\
%\text{s.t.} \quad & \textit{MAdds}(\cS) \leq q, \textit{Param}(\cS) \leq p \notag
%\end{align}

%-architecture search as an optimization problem

%- ma acc given param limit, flop limit

%-np hard

%-heuristics typically used for search (refs, darts, etc)

%% file: submod.tex
\subsection{Submodular Optimization}\label{sec:submod}
The search over the set of blocks that maximizes accuracy and remains within budget is NP-hard, even when the cost of each block with respect to number of parameters and computations is equal among all elements. However, using ideas from submodular optimization we can derive heuristics for architecture search.  
\begin{definition}\label{def:submod}
A function $F: 2^{\cV} \rightarrow \bR$, where $\cV$ is a finite set and $2^{\cV}$ denotes the power set of $\cV$, is \textbf{submodular} if for every $\cA \subseteq \cB \subseteq \cV$ and $v \in \cV \setminus \cB$ it holds that 
\begin{align}\label{eq:submod}
F(\cA \cup \{v\}) - F(\cA) \geq F(\cB \cup \{v\}) - F(\cB)
\end{align}
\end{definition}
Intuitively, submodular functions have a natural \textit{diminishing returns} property. Adding additional elements to an already large set is not as valuable as adding elements when the set is small. A subclass of submodular functions are \textbf{monotone}, where for any $\cA \subseteq \cB \subseteq \cV, F(\cA) \leq F(\cB)$. Submodular functions enjoy a number of other properties, including being closed under nonnegative linear combinations.

Typical submodular optimization $\max_{\cS\subseteq \cV} F(\cS)$ involves finding a subset $\cS\subseteq \cV$ given some constraints on the chosen set: cardinality constraints $|\cS|\leq r$ being the simplest. Formal optimization in these cases is NP-hard for general forms of submodular functions $F(\cS)$, and requires complex and problem-specific algorithms to find good solutions \cite{kawahara2009submodularity}. However, it has been shown that the \textit{greedy algorithm} can obtain good results in practice.

Starting with the empty set, the algorithm iteratively adds an element $v_k$ to the set $\cS_{k-1}$ with the update:
\begin{equation}\label{eq:gain}
v_k = \text{argmax}_{v\in \cV\setminus\cS_{k-1}} F(\cS_{k-1} \cup \{v\}) - F(\cS_{k-1}),
\end{equation}
% $\cS_{i+1} = \cS_{i} \cup \arg\max_v \Delta(v|S_{k})$, 
where $F(\cS_{k-1} \cup \{v\}) - F(\cS_{k-1})$ is the marginal improvement in $F$ of adding $v$ to the previous set $\cS_{k-1}$.
Results in \cite{krause2007near} show that for a nonnegative monotone submodular function, the greedy algorithm can find a set $\cS_r$ such that $F(\cS_r) \geq (1-1/e)\max_{|\cS|\leq r} F(\cS)$. We use this result to derive heuristics for finding good architectures.

%% file: arch.tex
\section{Submodular Neural Architecture Search}

%Figure \ref{??} presents an example for building a CNN with a selected subset of blocks. Each position in the network can be filled by a different basic block.
%This selection initiates from a very small subset of basic blocks to build CNN architecture for image classification and spreads over the all blocks of the given set.
%Varying accuracies may be achieved depending on which blocks are selected for each position in the ``sequence." Now we define the accuracy a given set of blocks achieves as a score set function $F: \cS \rightarrow \bR$.

%This cost is directly computed by the corresponding multiply-adds (MAdds) and number of parameters for every selected block, $c(\cS) = \sum_{v\in\cS} c(v)$. 

%Assume a sequence of $n$ blocks are to be picked for building an efficient CNN, and each block "position" has $l$ types that can be chosen from. Let $N$ denote the finite set of positions with $n$ elements (blocks), $N = \{1, \dots, n\}$, $L$ the set of all block types $L=\{1,\ldots,l\}$, and $F$ a set function defined over the power set of $N$, $F:l^N \rightarrow \bR$. Denote a block of type $j\in L$ at position $i\in N$ as $j_i$.

Assume $N$ block ``positions" need to be filled for building an efficient CNN, and each block {\em position} has $L$ types that can be chosen from.
%Let $N$ denote a finite set of $n$ positions $N = \{1, \dots, n\}$, $L$ the set of all block types $L=\{1,\ldots,l\}$,
Denote a block of type $l\in \cL=\{1,\ldots,L\}$ at position $n\in \cN=\{1, \dots, N\}$ as $l_n$,
$\cV$ denote a finite set with $N$ elements (blocks) 
%with $l$ types
and $F$ a set function defined over the power set of $\cV$, $F:L^\cV \rightarrow \bR$. 
% For example, $v=\{3_2\} \in l^\cV$
%If we directly follow the definition of submodularity (Def. \ref{def:submod}), the function $F$ is defined on an \textit{unordered} set. % and $F(S)=F(Perm(S))$. However, CNN architectures are sequential models, and this may not be the case in general.
%For example, given a set $S=\{1,4\}$, there are two sequences $S_1=\{1,4\}$ and $S_2=\{4,1\}$.
%Without considering the sequence order, $F(S_1) = F(S_2)$.
%Here, we assume function $F$ is defined on a \textit{sequence} in order to build a valid CNN architecture.
%For example, $S_2$ may not be a valid CNN architecture based on our design. We only consider the sequence $S_1$.
%(e.g., $\{4,1\}$ is not a valid architecture, so we take $\{1,4\}$).

Given a set of blocks, $\cS\in L^\cV$ (e.g., $\cS=\{3_2, 1_1\}$), we build the model CNN $cnn$ (e.g., the first block with type 1 ($1_1$) and the second with type 3 ($3_2$)), and take the validation accuracy of the CNN as the value of $F(\cS)$, $F(\cS)=f(cnn)$. % To be simple, we denote $F(\cS) = F(cnn)$, where the $cnn$ is built on the set with sequence order.
%In order to map a selected subset of blocks to a real value (accuracy), we assume a dataset set $X$ is given to train the constructed CNN from the selected blocks and evaluate its performance.
Then % with $F(\cS) = F(cnn)$, where $\cS\in l^N$, 
accuracy is exactly our map from the set of blocks to reals, and
%Define $F(\cS) = \frac{1}{m}\sum_{i=1}^{m}1_{h_{w_{\cS}}(x_i) = y_i}$, where $\cS \subseteq l^N$, instances $(\bm{x}_i, y_i), i=1,\dots m$ in $X_{\text{test}}$, $h$ is a cnn parameterized by $w_{\cS}$, $w_{\cS}$ is obtained by training the cnn $h$ on $X_{\text{train}}$ with cross entropy loss by stochastic gradient descent (sgd),
%\begin{equation}
%w_{\cS} = \arg \min_{w}-\sum_{i=1}^{k}y_i\log(h_w(\bm{x_i}))
%\end{equation}
%where $k$ is the mini-batch size. 
%Therefore
for each $\cS \in L^\cV$, a CNN is built with the selected blocks based on $\cS$. Our total search space size would be $L^N$. 
%Including locations where we have not inserted a block (an ``identity" block), our total search space size would be $(L+1)^N$. 

For each set of blocks $\cS$, the associated cost $c(\cS)$ cannot exceed the specified budget $B$. Using the above notions of accuracy and cost, our goal is to solve
the problem,
\begin{align}\label{eq:arch_opt}
	\underset{\cS \subseteq \cV}\max \quad F(\cS) \quad \text{subject to} \quad  c(\cS) \leq B 
\end{align}

%Consider the evaluation of $F(\cdot)$ at $\cS$. Using SGD training,
%we can optimize this well. But it is important to see that when
%we do this, for the overall architecture search,
%this simply provides us a (good evalution)
%value of $F(\cdot)$ at a single discrete position $\cS$. 
%Firstly, $F(\emptyset) = 0$, i.e., we do not improve the accuracy if we do not pick any blocks.
%In any case,
The accuracy objective $F(\cS)$ has an important property, given we can find the best possible parameter setting (global optimum) during SGD training.
$F$ is \textit{monotone}, i.e., $F(\cA) \leq F(\cB)$ for any $\cA \subseteq \cB \subseteq \cV$. Intuitively, adding blocks (making the network larger) can only
improve accuracy in general.
For each $\cS$, we can obtain its corresponding number of parameters $Param(\cS)$ or number of multiply-adds $MAdds(\cS)$. In practice,
training by SGD
may not reach the global optimum: in this case adding blocks may \textit{not} improve accuracy. However, our own empirical results and those in existing literature suggest
that this nondecreasing behavior is typically true, i.e., in ResNet \cite{he2016deep}.

Denote each cost-accuracy pair at global optimality as $(c_i, f_i), i=1, \dots, L^N$, and add three virtual points, $(0, 0)$, $(c_{L^N}, 0)$, $(c_{L^N}, \max \{f_1, \dots, f_{L^N}\})$.
This set can be seen as a \textit{convex hull}, where for each cost we assign its associated positive value on the convex hull. If $F(\cS)$ can always reach its convex hull point with respect to $c(\cS)$, the accuracy objective satisfies both nonnegativity and nondecreasing monotonicity.
This is exactly the diminishing returns property associated with submodularity: adding a block to a small set of selected blocks $\cA$ improves accuracy at least as much as if adding it to a larger selected block $\cB \supseteq \cA$. If we let the accuracy of the CNN be $0$ when no blocks are selected, $F(\emptyset) = 0$, then we immediately have that,
\begin{lemma}\label{thm:submodular}
	For any selected blocks $\cA \subseteq \cB \subseteq \cV$ and blocks $v \in \cV\setminus \cB$, it holds that 
	\begin{align}
	F(\cA) &\geq 0 \\
	F(\cA \cup \{v\}) - F(\cA) &\geq 0 \\
	F(\cA \cup \{v\}) - F(\cA) &\geq F(\cB \cup \{v\}) - F(\cB) 
	\end{align}
	where $F$ reaches its convex hull point w.r.t. the cost.
\end{lemma}
%A set function $F$ with diminishing return property is submodular. The proof of this theorem and other theorems can be seen in supplement.
Thus the neural architecture search problem can be
solved as
the problem of maximizing a nonnegative nondecreasing function, subject to parameter and computational budget constraints.

The simple greedy algorithm described in Section \ref{sec:submod} \eqref{eq:gain} assumes equal costs for all blocks. Naturally it can perform arbitrarily badly in the case where $c(\cS) = \sum_{v\in\cS} c(v)$, by iteratively adding blocks until the budget is exhausted. A block containing a very large number of parameters or expensive MAdds with accuracy $f_o$ will be preferred over a cheaper block offering accuracy $f_o - \epsilon$. To deal with these \textit{knapsack constraints}, the marginal gain update in \eqref{eq:gain} can be modified to the \textit{marginal gain ratio},
\begin{align}\label{eq:MGR}
	v_k &= \text{argmax}_{v\in \cV\setminus\cS_{k-1}} \frac{F(\cS_{k-1} \cup \{v\}) - F(\cS_{k-1})}{c(v)} %\\
%		v_k &= \text{argmax}_{e\in \cV\setminus\cS_{k-1}} \frac{F(\cS_{k-1} \cup v) - F(\cS_{k-1})}{MAdds(v)} \label{eq:AMR}
\end{align}
The modified greedy algorithm with marginal gain ratio rule attempts to maximize the cost$/$benefit ratio, and stops when the budget is exhausted. However, even with this modification, the greedy algorithm can still perform arbitrarily poorly with respect to global optima. For example, consider the parameter costs of picking between two blocks $v_1$ and $v_2$, Param($v_1$) = $\epsilon$, Param($v_2$) = $p$. If we compute the accuracy of adding the blocks as $F({v_1})=3\epsilon$, $F({v_2}) = 2p$, then the cost$/$benefit ratios are $\frac{F({v_1}) - F({\emptyset})}{Param(v_1)}=3$ and $\frac{F({v_2}) - F({\emptyset})}{Param(v_2)}=2$. The modified greedy algorithm will pick block $v_1$. If $v_1$ is picked and added to current set, and we do not have enough budget to next add $v_2$, we only achieve accuracy $3\epsilon$.  However, the optimal solution is to pick $v_2$ given any budget less than $p + \epsilon$. 

Fortunately, the greedy algorithm can be further adapted.
%to achieve a constant ratio factor approximation.
We compute $\tilde{\cS}_{APR}$ using the accuracy parameter ratio (APR) with rule \eqref{eq:MGR} and cost $Params(v)$, use the accuracy MAdds ratio (AMR) with the same  and cost $MAdds(v)$ to get $\tilde{\cS}_{AMR}$, \textit{and} take the \textit{uniform cost} (UC) greedy algorithm with rule Eq. \eqref{eq:gain} to get $\tilde{\cS}_{UC}$. The new modified Cost-Effective Greedy (CEG) algorithm returns the model which achieves maximum accuracy. With these rules, CEG can still achieve a constant ratio approximation.
\begin{theorem}\label{thm:CEG}
	If $F$ is a nondecreasing set function satisfying diminishing return property and $F(\emptyset)=0$, then the CEG algorithm achieves a constant ratio $\frac{1}{2}(1 - \frac{1}{e})$ of the optima:
	\begin{align}
	&\max\{F(\tilde{\cS}_{UC}), F({\tilde{\cS}_{APR}}), F(\tilde{\cS}_{AMR})\} \\ \notag
	&\geq
	\frac{1}{2}(1 - \frac{1}{e})\max_{MAdds(\cS) \leq B_m  Param(\cS) \leq B_p}F(\cS)
	\end{align}
\end{theorem}
%The Theorem \ref{thm:CEG} states that the best solution of $\tilde{\cS}_{pm}$, $\tilde{\cS}_{a/m}$ and  $\tilde{\cS}_{p/m}$ returned by CEG is at least a constant factor ratio $\frac{1}{2}(1 - \frac{1}{e})$ of the optimal solution. 
%While the theorem does not state that our architecture search problem is
%solved with quality guarantees, 
%it provides guidance that the general search scheme we are using,
%albeit in a heuristic way, is at least sensible. 
The proof is in the supplement. If we consider the time-cost of the accuracy function evaluation as O($T$) ($T$ is the time to train the network by SGD), then the running time of CEG is O($|\cV|\Phi T$), where $|\cV| = LN$ is the total number of blocks, $\Phi = \max_{1\leq k \leq LN} \{\frac{B_p}{Param(e_k)}, \frac{B_m}{MAdds(e_k)}\}$. 
The CEG algorithm is at most $O(T|\cV|^2)$. While this is a sizable improvement over our initial combinatorial approach, simple technical/empirical observations allow us to scale and speed up the CEG algorithm by early stopping during training and with lazy function evaluation. 

 {\footnotesize\begin{algorithm}[!t]
 		\caption{\label{alg:ceg} Cost-Effective Greedy CNN Search (CEG)}
 	{
 		\begin{algorithmic}
 			\Function {CEG}{$\cV$, $F$, $B$, $c(\cdot)$}
 			\State $\cS \leftarrow \arg\max_{v\in\cV} \frac{F(v)}{c(v)}$
 			\While {c($\cS$) $\leq$ $B$}
 			\State $v^{\star} = \arg\max_{v\in\cV\setminus\cS} \frac{F(S \cup \{v\}) - F(S)}{c(v)}$
 			\State $S \leftarrow S \cup \{v^{\star}\}$
 			\EndWhile
 			\State return $\cS$
 			\EndFunction
 		\end{algorithmic}}
\end{algorithm}}

\begin{figure}
	\centering
	\includegraphics[width=1.0\columnwidth]{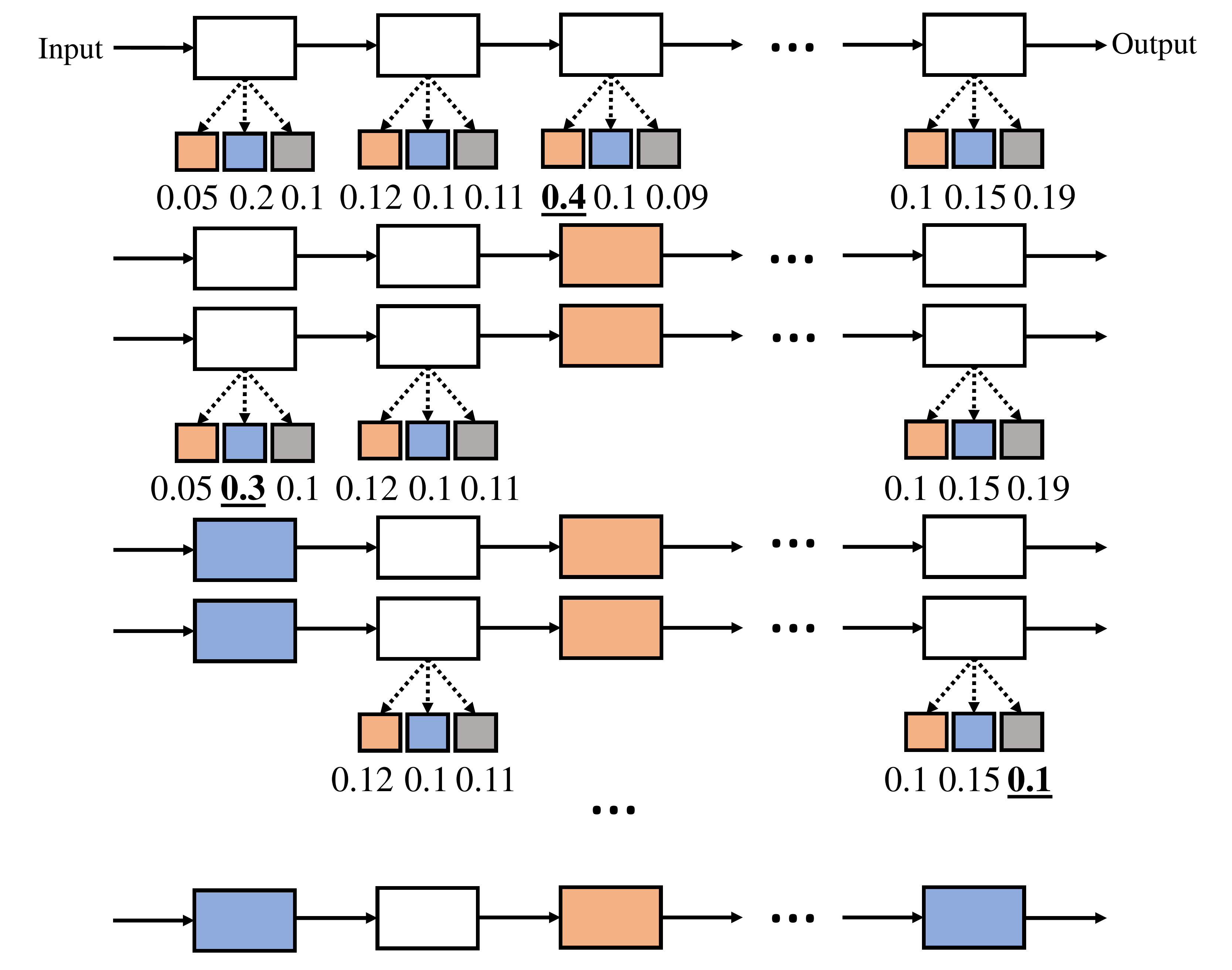}
	\caption{\label{fig:search}\footnotesize An overview of our Lazy Cost-Effective Greedy Search. Colors correspond to different basic block types, and empty boxes represent positions in the network to be considered for filling block. Numbers below blocks indicate marginal benefit of the block filled at that position. Step 1 searches over all blocks. Step 2 selects the block with highest marginal benefit, $1_3$, for filling in position 3. Step 3 updates the next highest marginal benefit with $1_3$ added. Step 4 picks the highest marginal benefit block, $2_1$, for filling in position 1. Step 5 updates the next highest marginal benefit with $\{1_3, 2_1\}$ added, $3_N$. However, the marginal benefit of $3_N$ is not the highest, so we did not pick this block and the search continues. The final architecture is obtained once the budget is exhausted (unfilled blocks are replaced with identity operations).}
	\vspace{-10pt}
\end{figure}

%-training accuracy is function

%-convex hull

%-simple greedy algorithm

%% file: speedup.tex
\subsection{Early Stop Training}
The running time is linear with respect to function evaluation time. This time includes the full training time to learn network weights by SGD in order to achieve test set accuracy close to the global optimum. This is expensive, and so even the ``low" number of O($|\cV|^2$) number of CNNs to search is prohibitive in practical situations in which one may not have access to large GPU clusters or months of training time. In this case, to reduce the time to learn the weights of CNNs for accuracy function evaluation, we reduce the number of epochs and train the network with early stopping. Our experiments indicate this leads to learned architectures similar to those when trained to optimality.

%(This could probably be a second paragraph here, also i dont understand it)
%{\color{red}Moreover, when CEG tries to add the next block, we cache the network we explored and it allows us to do fast lookup of the network without training it repeatedly.} 

%% file: reduce.tex
\subsection{Lazy Function Evaluation}
With early stopping and caching, while we can quickly evaluate the accuracy function $F(\cS)$ when adding any block, we still need to make a large number of function evaluations in order to run CEG. The running time is at least linear in the total number of blocks, and in the worst case quadratic if the number of selected blocks can be as high as total number of blocks $LN$. If we select $K$ blocks for building our CNN among $|\cV|$ blocks, O($K|\cV|$) evaluations are needed. The submodularity property can be exploited further to perform far fewer function evaluations.

The marginal benefit of adding any one block can be written as, 
for all $v\in \cV \setminus \cS$, $\Delta(v|\cS) = F(\cS \cup \{v\}) - F(\cS)$. The key idea of exploiting submodularity is that, as the set of our selected blocks grows, the marginal benefit can never increase, which means that if
$\cA \subseteq \cB \subseteq \cV$, it holds that $\Delta(v|\cA) \geq \Delta(v|\cB)$.
Therefore we do not need to update $\Delta(v|\cS)$ for every network after adding a new block $v^{\prime}$ and we can perform \textit{lazy evaluation}. This Lazy Cost Effective Greedy algorithm (LCEG), Alg. \ref{alg:lceg}, can be described as follows:
1. Apply rule Eq. \ref{eq:gain} to search all $LN$ blocks and keep an ordered list of all marginal benefits with decreasing order by priority queue. 2. Select the top element of the priority queue as the first selected block at the first iteration. 3. Reevaluate $\Delta(v|\cS)$ for the top element $v$ in the priority queue. 4. If adding block $v$, $\Delta(v|\cS)$ is larger than the top element in priority queue, so pick block $v$. Otherwise, insert it with the updated $\Delta(v|\cS)$ back in the queue. 5. Repeat steps 2-4 until the budget is exhausted. An overview of LCEG can be seen in Figure \ref{fig:search}.

In many cases, the reevaluation of $\Delta(v|\cS)$ will result in a new but not much smaller value, and the top element will often stay at the top. In this way, we can often find the next block to add without having to recompute all blocks. This lazy evaluation thus leads to far fewer evaluations of $F$ and means we need to train much fewer networks when trying to add one block. The final algorithm includes taking advantage of our LCEG procedure. Resource Constrained Architecture Search, RCAS is defined in Algorithm \ref{alg:lcas}.
% With respect to benefit cost ratio, for all $v\in \cV \setminus \cS$, 
%\begin{align}
%	\frac{\Delta(v|\cS)}{Param(v)} &= \frac{F(\cS \cup v) - F(\cS)}{Param(v)} \\
%		\frac{\Delta(v|\cS)}{MAdds(v)} &= \frac{F(\cS \cup v) - F(\cS)}{MAdds(v)}
%\end{align}
%we also apply this lazy evaluation procedure for reducing the number of networks to train. Lazy cost-efficient greedy algorithm (LCEG) is provided in Algorithm \ref{alg:lceg} and can be implemented very easily.

{\footnotesize\begin{algorithm}[!t]
		\caption{ \label{alg:lceg} Lazy Cost-Effective Greedy Search}
		\begin{algorithmic}
			\Function {Lazy-CEG}{$\cV$, $F$, $B_p$, $B_m$, $c(\cdot)$}
			\State $\cS \leftarrow \emptyset$
			\State $PriorityQueue$ $Q \leftarrow PriorityQueue()$
			\ForAll {$v \in \cV$}
			\Comment{First iteration}
			\If {Param($v$) $\leq$ $B_p$ AND MAdds($v$) $\leq$ $B_m$}
			%			\State Q.push($F(S \cup e) - F(S)$)
			%			\EndIf
			%			\If {type == $p/m$}
			\State Q.push($\{v, \frac{F(v)}{c(v)}\}$)
			\EndIf
			%			\If {type == $a/m$}
			%			\State Q.push($\frac{F(S \cup e) - F(S)}{MAdds(e)}$)
			%			\EndIf
			
			\EndFor
			%\State $v^{\star} \leftarrow Q.pop()$
			\State $S \leftarrow S \cup \{Q.pop()\}$
			\While {$\exists v \in Q$ :Param($\cS \cup \{v\}$) $\leq$ $B_p$ AND MAdds($\cS \cup \{v\}$) $\leq$ $B_m$}
			\Comment{Lazy update}
			\State $v^{\star} \leftarrow Q.pop()$
			\If {$v^{\star} \in \cV \setminus \cS$}
			%\If {type == pm}
			%			\If {$F(S \cup e) - F(S) \geq F(S\cup Q.top()) - F(S)$}
			%			\State $S \leftarrow S \cup e^{\star}$
			%			\Else
			%			\State Q.push($e^{\star}$)
			%			\EndIf
			%			
			%			\ElsIf {type == $p/m$}
			%			\If {$\frac{F(S \cup e^{\star}) - F(S)}{Param(e^{\star})} \geq \frac{F(S\cup Q.top()) - F(S)}{Param(e)}$}
			%			\State $S \leftarrow S \cup e^{\star}$
			%			\Else
			%			\State Q.push($e^{\star}$)
			%			\EndIf
			
			%			\Else
			\If {$\frac{F(S \cup \{v^{\star}\}) - F(S)}{c(v^{\star})} \geq \frac{F(S\cup \{Q.top()\}) - F(S)}{c(v)}$}
			\State $S \leftarrow S \cup \{v^{\star}\}$
			\Else
			\State Q.push($\{v^{\star}, \frac{F(S \cup \{v^{\star}\}) - F(S)}{c(v^{\star})} \}$)
			\EndIf
			\EndIf
			%			\EndIf

			\EndWhile
			\State return $\cS$
			\EndFunction
		\end{algorithmic}
	\end{algorithm}}

 {\footnotesize\begin{algorithm}[!t]
 		\caption{ \label{alg:lcas} Resource Constrained Architecture Search (RCAS)}
 		\begin{algorithmic}			
			
			\Function {\textbf{RCAS}}{$\cV$, $F$, $B_p$, $B_m$}
			\State $\tilde{\cS}_{UC} \leftarrow $ \Call{Lazy-CEG}{$\cV$, $F$, $B_p$, $B_m$, $const(\cdot)$} 
			\State $\tilde{\cS}_{APR} \leftarrow $ \Call{Lazy-CEG}{$\cV$, $F$, $B_p$, $B_m$, Param($\cdot$)} 
			\State $\tilde{\cS}_{AMR} \leftarrow $ \Call{Lazy-CEG}{$\cV$, $F$, $B_p$, $B_m$, MAdds($\cdot$)} 
			\State return $\arg\max\{\tilde{\cS}_{UC}, \tilde{\cS}_{APR}, \tilde{\cS}_{AMR}\}$
			\EndFunction
		\end{algorithmic}
\end{algorithm}}

\begin{table*}[]
	\centering
	{\footnotesize \begin{tabular} {c|ccc}
			\hline
			Layer & Input & Operator & Output\\
			\hline\hline
			Group-wise expansion layer & $H \times W \times C_1$ & 1x1 gconv2d group$=g_e$, ReLU6 & $H \times W \times (C_1 \times t)$ \\
			\hline
			Depthwise layer & $H \times W \times (C_1 \times t)$ & 3x3 dwise stride = $s$, ReLU6
			& $H/s\times W/s\times (C_1 \times t)$ \\
			\hline
			Group-wise projection layer & $H/s\times W/s\times (C_1 \times t)$ & linear 1x1 gconv2d group = $g_p$ & $H/s\times W/s\times C_2$ \\
			\hline
		\end{tabular}
	}
	\vspace{-7pt}
	\caption{\label{tab:blocks} \footnotesize Parameter and performance efficient depth-wise based basic blocks used in Resource Constrained Architecture Search. The structure of basic blocks derive from depth-wise based MobileNetV2 blocks, changing the expansion factor $t$ and using group convolutions. Basic blocks transform from $C_1$ to $C_2$ channels with expansion factor $t$, expansion group $g_e$ and projection group $g_p$ with stride $s$.  }
	\vspace{-5pt}
\end{table*}

%% file: exps.tex
\section{Experiments}
It is expensive to search directly for CNN models on ImageNet and it can take several days to find a network architecture (even without resource constraints). Previous works \cite{zoph2017learning,real2018regularized} suggest that we can perform our architecture search experiments on a smaller proxy task and then transfer the top-performing architecture discovered during search to the target task. However, \cite{tan2018mnasnet} shows that it is not-trivial to find a good proxy task under constraints. Experiments on CIFAR-10 \cite{krizhevsky2009learning} and the Standford Dogs Dataset\cite{khosla2011novel} demonstrate these datasets are not good proxy tasks for ImageNet when a budget constraint is taken into account. RCAS shines in this problem setting, allowing us to perform our architecture search on a much larger dataset, CIFAR-100. Indeed, we also can directly perform our architecture search on the ImageNet training set, to directly evaluate and compare the architectures learned.
In these large scale cases, we train for fewer steps on CIFAR-100 and ImageNet.

\begin{table}[!b]
	\centering
	\vspace{-10pt}
	\begin{center}
		\begin{tabular}{ccc}
			\includegraphics[width=0.3\columnwidth]{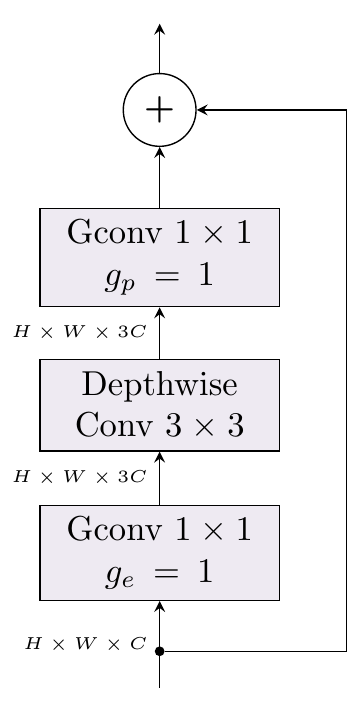} &
			\includegraphics[width=0.3\columnwidth]{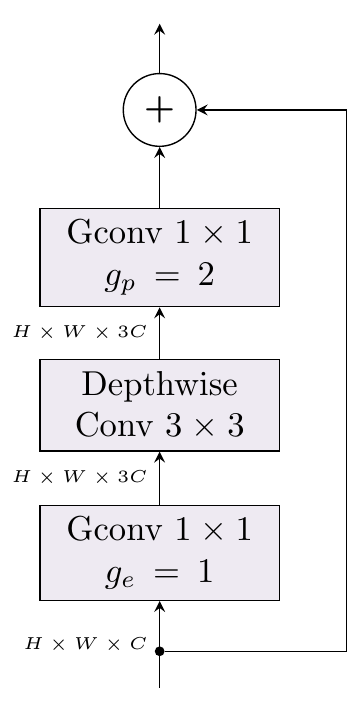} &
			\includegraphics[width=0.3\columnwidth]{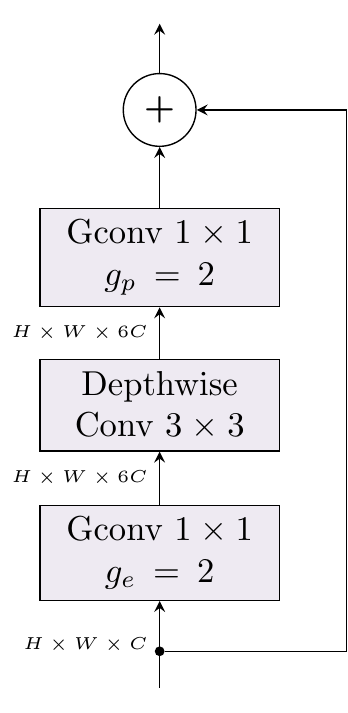} \\
			\small Type 1 & \small Type 2 & \small Type 3\\
			\includegraphics[width=0.3\columnwidth]{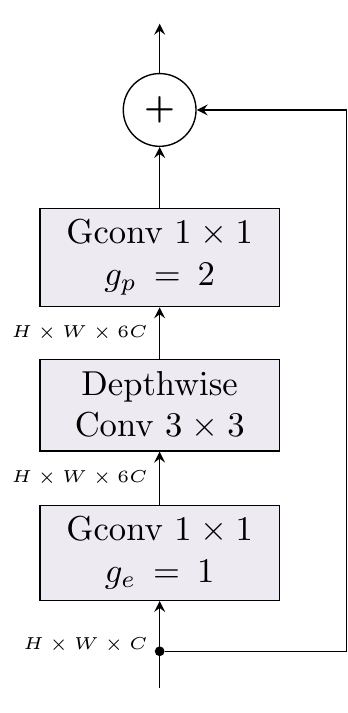} &
			\includegraphics[width=0.3\columnwidth]{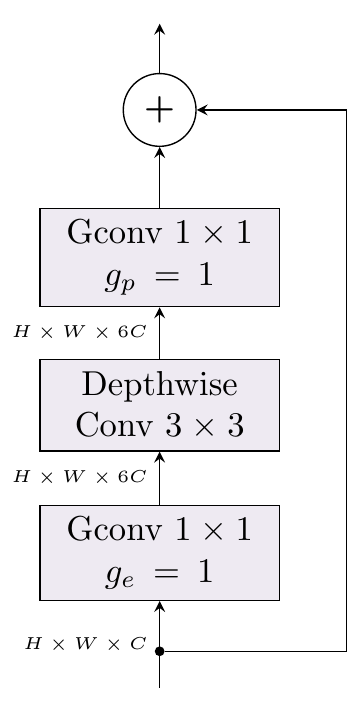} &
			\includegraphics[width=0.3\columnwidth]{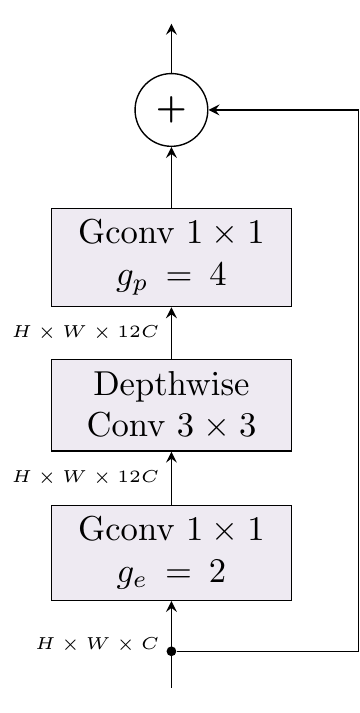} \\
			\small Type 4 & \small Type 5 & \small Type 6\\
		\end{tabular}
		\vspace{-7pt}
		\captionof{figure}{\footnotesize\label{tab:types} The 6 types depthwise-based basic blocks used in RCAS.}
	\end{center}
\end{table}

Our experiments on CIFAR-100 and ImageNet have two steps: architecture search and architecture evaluation. In the first step, we search for block architectures using RCAS and pick the best blocks based on their validation performance. In the second step, the picked blocks are used to build CNN models, which we train from scratch and evaluate the performance on the test set. Finally, we extend the network architecture learned from CIFAR-100 and evaluate the performance on ImageNet, comparing with the architecture learned through RCAS applied directly to ImageNet.

%\begin{figure}\label{fig:blocks}
%	\centering
%	\begin{subfigure}[b][width=0.3\columnwidth]
%		\includegraphics[]{./figs/basic_blocks/type_1.pdf}
%		\caption{$g_e=1, t = 3, g_p=1$}
%	\end{subfigure}
%	\begin{subfigure}[b]
%		\includegraphics[width=0.3\columnwidth]{figs/basic_blocks/type_2.pdf}
%		\caption{$g_e=1, t = 3, g_p=2$}
%	\end{subfigure}
%	\begin{subfigure}[b]
%	\includegraphics[width=0.3\columnwidth]{figs/basic_blocks/type_3.pdf}
%	\caption{$g_e=2, t = 6, g_p=2$}
%\end{subfigure}
%	\begin{subfigure}[b]
%	\includegraphics[width=0.3\columnwidth]{figs/basic_blocks/type_4.pdf}
%	\caption{$g_e=1, t = 6, g_p=2$}
%\end{subfigure}
%	\begin{subfigure}[b]
%	\includegraphics[width=0.3\columnwidth]{figs/basic_blocks/type_5.pdf}
%	\caption{$g_e=1, t = 6, g_p=1$}
%\end{subfigure}
%	\begin{subfigure}[b]
%	\includegraphics[width=0.3\columnwidth]{figs/basic_blocks/type_6.pdf}
%	\caption{$g_e=2, t = 12, g_p=4$}
%\end{subfigure}
%	\caption{The 6 basic blocks used in LCEG architecture search.}
%\end{figure}
\begin{table*}[]\centering
	\resizebox{2.0\columnwidth}{!}{
		\begin{tabular}{c|c c c c c c} 
			\hline
			Architecture & Top-1 Accuracy & Top-5 Accuracy & Parameters & MAdds & Search Method & Cost (GPU Days) \\ [0.5ex]
			\hline\hline
			MobileNetv2 & 74.2 & 93.3 & 2.4M & 91.1M & manual & - \\ 
			ShuffleNet(1.5) & 70.0 & 90.8 & 2.3M & 91.0M & manual & - \\ 
			NASNet-A & 70.0 & 86.0 & 3.61M & 132.0M & RL & 1800 \\ 
			DARTS (searched on \textbf{CIFAR-10}) & 75.9 & 93.8 & 3.40M & 198.0M & gradient-based & 4\\ 
			\hline
			RCNet (searched on \textbf{CIFAR-100}) & \textbf{76.1} & \textbf{94.0} & \textbf{1.92M} & \textbf{87.3M} & RCAS & \textbf{2} \\
			\hline
	\end{tabular}}
	\vspace{-7pt}
	\caption{\footnotesize \label{tab:cifar100accu}Comparison with state-of-the-art image classifiers on CIFAR-100. Our searched model performs significantly better than other manual methods. Given MobileNetV2 parameter and MAdds constraints, our model still outperforms DARTS with $\sim44\%$ fewer parameters and $\sim50\%$ fewer MAdds. Additionally, both RCAS and RCNet run on CIFAR-100 much faster than DARTS. }
	\vspace{-10pt}
\end{table*}

\subsection{Architecture Search}
As our main purpose is to look for low cost mobile neural networks, the following basic blocks (using depthwise convolution extensively) are included for architecture search, varying from MobileNetV2 blocks by using different expansion ratios and group convolutions for expansion and projection (see Table \ref{tab:blocks}). Each type of block is shown in Figure \ref{tab:types} and consists of different types of layers. We have $L = 6$ different basic blocks to pick from and $N = 36$ number of positions can be filled for building networks under parameter and MAdds constraints for our low cost architecture search. An overview of picking basic blocks to fill positions can be seen in Figure \ref{fig:search}. During architecture search, only one basic block can be picked to fill a position, otherwise the procedure will not insert any block. The input for the $k^{th}$ picked block is the output of the $(k - 1)^{th}$ picked block, naturally stacking and forming the network.

Standard practice in architecture search \cite{real2018regularized} suggests a separate validation set be used to measure accuracy: we randomly select 10 images per class from the training set as the fixed validation set. During architecture search by RCAS, we train each possible model (adding one basic block with different options at every position) on 10 epochs of the proxy training set using an aggressive learning rate schedule, evaluating the model through the accuracy set  function $F(\mathcal{S})$ on the fixed validation set. We use stocastic gradient descent (SGD) to train each possible model for training with \textit{nesterov} momentum set to 0.9. We employ a multistep learning rate schedule with initial learning rate $0.1$ and multiplicative decay rate $g = 0.1$ at epochs
$4, 7 $ and $9$ for fast learning. We set the regularization parameter for weight decay to $4.0e^{-5}$, following InceptionNet \cite{szegedy2017inception}. 

\begin{figure}[!b]\centering
	\vspace{-15pt}
	\includegraphics[width=.9\columnwidth]{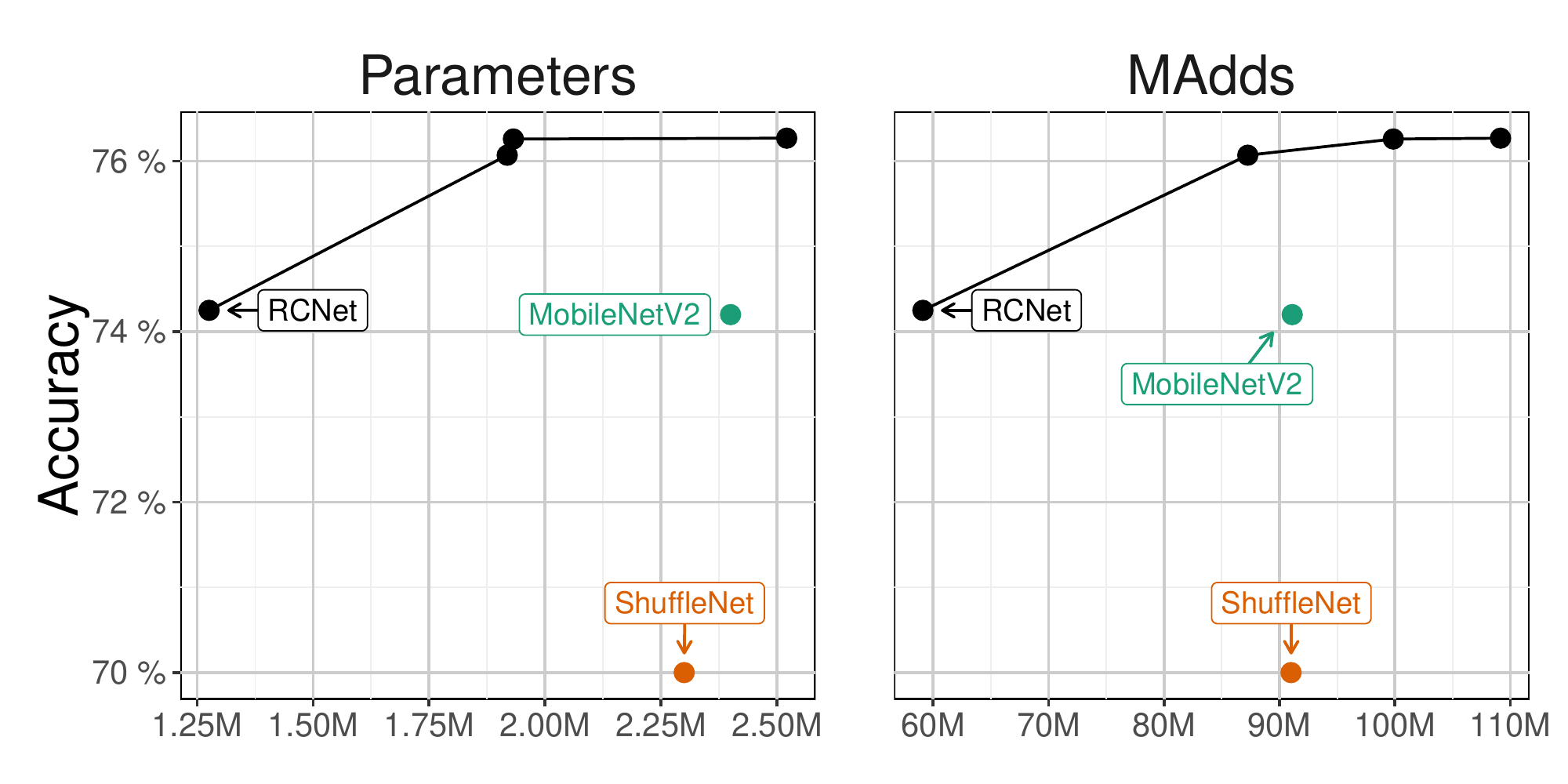}
	\vspace{-15pt}
	\caption{\label{fig:sprocess}\footnotesize Architectures along the search path over constraints are stored. RCAS uses ($\sim 45\%$) fewer parameters and ($\sim 35\%$) fewer MAdds to achieve similar accuracy. }
\end{figure}

\subsection{Architecture Evaluation}
Applying our RCAS, we obtain a selected architecture under the given parameter and MAdds budget. To evaluate the selected architecture, we train it from scratch and evaluate the computational efficiency and accuracy on the test set. Given mobile budget constraints, we compare our selected architecture with mobile baselines, MobileNetV2 and ShuffleNet \cite{sandler2018mobilenetv2,hluchyj1991shuffle}. We take the parameter number and MAdds as the computational efficiency and report the latency and model size on a typical mobile platform (iPhone 5s). We evaluate the performance of our selected architecture on both the CIFAR-100 dataset and ImageNet dataset. Following prior work \cite{szegedy2017inception}, we use the validation dataset as a proxy for test set ImageNet classification accuracy.
Our RCAS algorithm and subsequent Resource Constrained CNN (RCNet) are implemented using PyTorch\cite{paszke2017pytorch}. We use built-in $1\times 1$ convolution and group convolution implementations. The methods are easy to reproduce in other deep learning frameworks such as Caffe \cite{jia2014caffe} and TensorFlow\cite{abadi2016tensorflow}, using built-in layers as long as $1\times 1$ standard convolutions and group convolutions are available. For CIFAR-100, we use similar parameter settings as during search, with the exception of a maximum number of epochs of 200 and a learning rate schedule updating at epochs 60, 120, and 180. For ImageNet, we use an initial learning rate 0.01, and decay at epochs 200 and 300 with maximum training epoch 400. 
We use the same default data augmentation module as in ResNet for fair comparisons.
Random cropping and horizontal flipping are used for training
images, and images are resized or cropped to $224\times 224$
pixels for ImageNet and $32\times 32$ pixels for CIFAR-100.
At test time, the trained model is evaluated on center crops.
%The same default settings are used in image preprocessing. 

\subsection{CIFAR-100}
The CIFAR-100 dataset \cite{krizhevsky2009learning} consists of $50,000$ training RGB images and $10,000$ test RGB images with $100$ classes. The image size is $32\times 32$. We take the state-of-the-art mobile network architecture MobileNetV2 as our baseline. All the hyperparameters and preprocessing are set to be the same in order to make a fair comparison. The $32\times 32$ images are converted to $40\times 40$ with zero-padding by 4 pixels and then randomly cropped to $32\times 32$. Horizontal flipping and RGB mean value substraction are applied as well. 
\begin{figure}[!t]\centering
	\vspace{-5pt}
	\includegraphics[width=0.9\columnwidth]{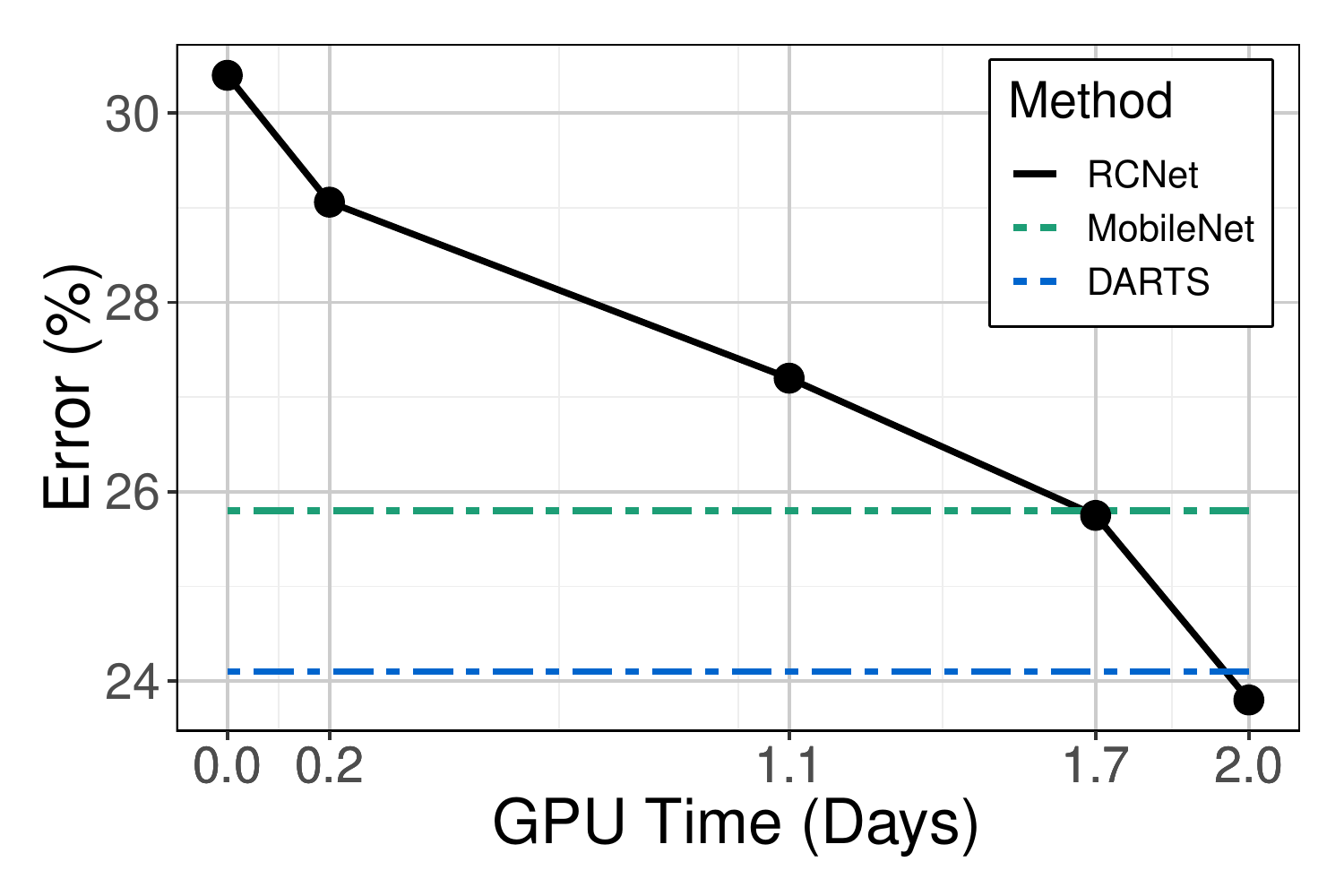}
	\vspace{-10pt}
	\caption{\label{fig:mobilecmp}\footnotesize Search progress of RCAS on CIFAR-100. We keep track of the most recent architecture over time, with MobileNetV2 and the final DARTS architecture as reference.}
	\vspace{-10pt}
\end{figure}
%The overall LCNet architecture for CIFAR100 are the same as ImageNet in Table \ref{??} except for different input and output size and strides of the first conv2d. 

%As our purpose is to build resource bounded image classifier on mobile platform, we only compare our model with low computational cost models with fewer parameters
%consuming less memory and taking small network width.
%We consider mobile-suitable models, MobileNet and ShuffleNet
%as our comparison baselines.
We evaluate the top-1
and top-5 accuracy and compare MAdds and the number of
parameters for benchmarking. The performance comparison
between baseline models and our RCNet is shown in Table \ref{tab:cifar100accu}. RCNet achieves significant
improvements over MobileNetV2 and ShuffleNet
with fewer computations and fewer parameters. Our
RCNet achieves similar accuracy with MobileNetV2 with $\sim 45\%$ parameter reduction and $\sim 35\%$ computation reduction (see Figure \ref{fig:sprocess}). With $\sim 20\%$ fewer parameters, RCNet achieves a $1.9\%$ accuracy improvement. Search progress on CIFAR-100 can be seen in Figure \ref{fig:mobilecmp}.

\textbf{Does submodular property with early stopping hold?} On CIFAR-100, we find that our procedure displays monotonicity and diminishing returns over accuracy averaged over 500 networks built on random block additions (out of our 6 types) with early stopping:  56.96\%, 57.98\%, 58.81\%, 59.47\%, 59.94\%, 60.35\%, 60.70\%, 60.72\%. It may be the case that with early stopping we may not be identifying
the absolute best block at a given step in the algorithm, but as demonstrated, empirically we find that the final architecture identified is competitive with state-of-the-art.

\textbf{Does block diversity help?} To show the gains from block diversity, we use our method to search \textit{only} with block type 5. On CIFAR-100, we obtain a top-1 accuracy of $75.8\%$, worse than the original searched model with $76.1\%$, under the same constraints. This points to the importance of block diversity in resource constrained CNN models. 

\textbf{Does our search procedure help?} To show the gains from our search procedure, we use the searched solution and replace the last several blocks with
random blocks. With $\sim 1.8$M parameters and $\sim 73.0$M 
MAdds, the random-block solution only yields $74.9\%$ top-1 accuracy on CIFAR-100, as opposed to the original searched solution, $76.1\%$. Randomly adding one more block only gives $75.1\%$ top-1 accuracy with $\sim 2.9$M parameters and $\sim 88$M MAdds.
\begin{table*}[]\centering
	\resizebox{2.0\columnwidth}{!}{
		\begin{tabular}{c|c c c c c c} 
			\hline
			Architecture & Top-1 Accuracy & Top-5 Accuracy & Parameters & MAdds & Search Method & Cost (GPU Days) \\ [0.5ex]
			\hline\hline
			InceptionV1\cite{szegedy2015going} & 69.8 & 89.9 & 6.6M & 1448M & manual & - \\ 
			MobileNetV1\cite{howard2017mobilenets} & 70.6 & 88.2 & 4.2M & 575M & manual & - \\
			ShuffleNet(1.5)\cite{Zhang_2018_CVPR} & 71.5 & - & 3.4M & 292M & manual & - \\
			CondenseNet(G=C=4)\cite{huang2017densely} & 71.0 & 90.0 & 2.9M & 274M & manual & - \\
			MobileNetV2\cite{sandler2018mobilenetv2} & 72.0 & 91.0 & 3.4M & 300M & manual & - \\
			ANTNet\cite{xiong2019antnets} & 73.2 & 91.2 & 3.7M & 322M & manual & - \\
			\hline 
			NASNet-A\cite{zoph2017learning} & 74.0 & 91.6 & 5.3M & 564M & RL & 1800 \\ 
			AmoebaNet-A\cite{real2018regularized} & 74.5 & 92 & 5.1M & 555M & RL & 1800 \\ 
			MNasNet-92 (searched on \textbf{ImageNet})\cite{tan2018mnasnet} & \textbf{74.8} & \textbf{92.1} & 4.4M & 388M & RL & - \\
			Proxyless-R\cite{cai2018proxylessnas} & 74.6 & 92.2 & - & - & RL & 9 \\
			PNASNet\cite{liu2018progressive} & 74.2 & 91.9 & 5.1M & 588M & SMBO & $\sim$255 \\
			DPP-Net-Panaca (searched on \textbf{CIFAR-10})\cite{dong2018dpp}& 74.0 & 91.8 & 4.8M & 523M & SMBO & \textbf{8}$^{\dagger}$ \\ 
			DARTS (searched on \textbf{CIFAR-10})\cite{liu2018darts}& 73.1 & 91.0 & 4.9M & 595M & gradient-based  & \textbf{4}$^{\dagger}$ \\ 
			FBNet-C\cite{wu2019fbnet} & 74.9 & - & 5.5M & 375M & gradient-based & 9 \\
			\hline
			RCNet (searched on \textbf{ImageNet}) & 72.2 & 91.0 & 3.4M & 294M & RCAS & \textbf{8} \\
			RCNet-B (searched on \textbf{ImageNet}) & \textbf{74.7} & \textbf{92.0} & 4.7M & 471M & RCAS & \textbf{9} \\
			\hline
	\end{tabular}}
	\vspace{-7pt}
	\caption{\label{tab:imagenet}\footnotesize Performance Results on ImageNet Classification. Given $3.4M$ parameters and $300M$ MAdds constraints, RCAS finds a model searching on ImageNet using 8 GPU days, much faster than other automated methods and RCNet performs better than ``manual" methods with similar complexity. With $5M$ parameters and $500M$ MAdds constraints, RCNet-B achieves comparable accuracy to MNasNet-92 with fewer computation resources. RCNet-B outperforms DPP-Net-Panaca by $0.7\%$ and DARTS by $1.6\%$ with similar computation resources (The methods marked by $\dagger$ are searched on CIFAR-10, while our method is searched on ImageNet directly).}
	\vspace{-10pt}
\end{table*}
\subsection{ImageNet}
There are $1.28M$ training images and $50K$ validation images from $1,000$ classes in the ImageNet dataset\cite{deng2009imagenet}. Following the procedure for CIFAR-100, we learn the RCNet architecture on the training set and report top-1 and top-5 validation accuracy with the corresponding parameters and MAdds of the model. The details of our learned RCNet architecture can be seen in the supplement. We compare our models with other low cost models (e.g. $\sim 3.4M$ parameters and $\sim 300M$ MAdds) in Table \ref{tab:imagenet}. RCNet achieves consistent improvement over MobileNetV2 by $0.2\%$ Top-1 accuracy and ShuffleNet ($1.5\%$) by $0.7\%$. Compared with the most resource-efficient model, 
CondenseNet ($G=C=4$), our RCNet performs better with $1.2\%$ accuracy gain. 

Using the model found with CIFAR-100, we retrain the same model with ImageNet. Performance is comparable to MobileNetV2 with similar complexity, indicating that our procedure can effectively transfer to new and challenging datasets. Here, the adapted RCNet obtains favorable results compared with state-of-the-art RL search methods, with three orders of magnitude fewer computational resources. More details can be found in the supplement.
The final model constructed by RCAS includes 18 basic blocks with 6 types in the following sequence:
\vspace{-5pt}
\begin{align*}
[5, 1, 4, 5, 1, 5, 2, 4, 1, 6, 4, 6, 5, 3, 3, 6, 3, 6]
\end{align*}
\begin{table}[!b]
	\vspace{-5pt}
	\centering
	\resizebox{\columnwidth}{!}{
		\begin{tabular} {c|ccc}
			\hline
			Model &  MAdds & CoreML Model Size & Inference Time\\
			\hline\hline
			MobileNetV2 &  $300$ M & $14.7$ MB & $197.2$ ms \\
			RCNet & $294$ M & $14.6$ MB & $183.5$ ms \\
			\hline
		\end{tabular}
	}
	\vspace{-8pt}
	\caption{\label{tab:iphone5s}\footnotesize Inference time running on an actual device, iPhone 5s. As expected, our searched model, RCNet, use similar average inference time as MobileNetV2 per image.}
	%\vspace{-10pt}
\end{table}
{\bf Remarks.} There are a few interesting observations to be made here. First, given the limited parameter and MAdds budget, RCAS picks very few blocks with higher cost. Additionally, picking too many high dimensional blocks decreases the performance of the model compared to selecting fewer more low dimensional blocks. Additional details regarding the search path is in the supplement.
Second, with a specified maximum cost set to approximately the size of MobileNetV2, we identify a similar number of blocks. However, the blocks identified are \textit{diverse}.
Common mobile architectures consist of replications of the same type of block, e.g., MobileNetV2. This may suggest that block diversity is a valuable component of designing resource constrained mobile neural networks.

\subsection{Inference Time}
We test the actual inference speed on an iOS-based phone, iPhone 5s (1.3 GHz dual-core Apple A7 processor and 1GB RAM), and compare with baseline model MobileNetV2.
%The device has a .
To run the models, we convert our trained model to a \textit{CoreML} model and deploy it using Apple's machine learning platform. We report the inference time of our models in Table \ref{tab:iphone5s} (average over 10 runs). As expected, RCNet and MobileNetV2 have similar inference times.

%% file: conclusion.tex
\section{Conclusion}
Mobile architecture search is becoming an important topic
% with computer
in computer vision,
where algorithms are increasingly being integrated and deployed
%vision methods becoming increasingly integrated within applications
%and deployed
on heterogenous small devices. 
Borrowing ideas from submodularity, we propose algorithms for resource constrained 
architecture search. With resource constraints defined by model size and complexity,
we show that we can efficiently search for neural network architectures that perform quite well.
On CIFAR-100 and ImageNet, we identify mobile architectures that match or
outperform existing methods, but with far fewer parameters and computations.
Our algorithms are easy to implement and can be directly extended to identify
efficient network architectures in other resource-constrained applications. Code/supplement is available at \href{https://github.com/yyxiongzju/RCNet}{https://github.com/yyxiongzju/RCNet}.

\noindent {\bf Acknowledgments.}
This work was supported by  NSF CAREER award RI 1252725, UW CPCP (U54AI117924) and a NIH predoctoral fellowship to RM via T32 LM012413. We thank Jen Birstler for their help with figures and plots, and Karu Sankaralingam
for introducing us to this topic. 

%% file: main_camera.bbl
\begin{thebibliography}{10}\itemsep=-1pt

\bibitem{abadi2016tensorflow}
Mart{\'\i}n Abadi, Paul Barham, Jianmin Chen, Zhifeng Chen, Andy Davis, Jeffrey
  Dean, Matthieu Devin, Sanjay Ghemawat, Geoffrey Irving, Michael Isard, et~al.
\newblock Tensorflow: A system for large-scale machine learning.
\newblock In {\em 12th $\{$USENIX$\}$ Symposium on Operating Systems Design and
  Implementation ($\{$OSDI$\}$ 16)}, pages 265--283, 2016.

\bibitem{ashok2017n2n}
Anubhav Ashok, Nicholas Rhinehart, Fares Beainy, and Kris~M Kitani.
\newblock N2n learning: Network to network compression via policy gradient
  reinforcement learning.
\newblock {\em arXiv preprint arXiv:1709.06030}, 2017.

\bibitem{baker2016designing}
Bowen Baker, Otkrist Gupta, Nikhil Naik, and Ramesh Raskar.
\newblock Designing neural network architectures using reinforcement learning.
\newblock {\em arXiv preprint arXiv:1611.02167}, 2016.

\bibitem{brock2017smash}
Andrew Brock, Theodore Lim, James~M Ritchie, and Nick Weston.
\newblock Smash: one-shot model architecture search through hypernetworks.
\newblock {\em arXiv preprint arXiv:1708.05344}, 2017.

\bibitem{cai2018efficient}
Han Cai, Tianyao Chen, Weinan Zhang, Yong Yu, and Jun Wang.
\newblock Efficient architecture search by network transformation.
\newblock AAAI, 2018.

\bibitem{cai2018proxylessnas}
Han Cai, Ligeng Zhu, and Song Han.
\newblock Proxylessnas: Direct neural architecture search on target task and
  hardware.
\newblock {\em arXiv preprint arXiv:1812.00332}, 2018.

\bibitem{deng2009imagenet}
Jia Deng, Wei Dong, Richard Socher, Li-Jia Li, Kai Li, and Li Fei-Fei.
\newblock Imagenet: A large-scale hierarchical image database.
\newblock In {\em 2009 IEEE conference on computer vision and pattern
  recognition}, pages 248--255. Ieee, 2009.

\bibitem{dong2018dpp}
Jin-Dong Dong, An-Chieh Cheng, Da-Cheng Juan, Wei Wei, and Min Sun.
\newblock Dpp-net: Device-aware progressive search for pareto-optimal neural
  architectures.
\newblock In {\em Proceedings of the European Conference on Computer Vision
  (ECCV)}, pages 517--531, 2018.

\bibitem{gordon2018morphnet}
Ariel Gordon, Elad Eban, Ofir Nachum, Bo Chen, Hao Wu, Tien-Ju Yang, and Edward
  Choi.
\newblock Morphnet: Fast \& simple resource-constrained structure learning of
  deep networks.

\bibitem{han2015learning}
Song Han, Jeff Pool, John Tran, and William Dally.
\newblock Learning both weights and connections for efficient neural network.
\newblock In {\em Advances in neural information processing systems}, pages
  1135--1143, 2015.

\bibitem{he2016deep}
Kaiming He, Xiangyu Zhang, Shaoqing Ren, and Jian Sun.
\newblock Deep residual learning for image recognition.
\newblock In {\em Proceedings of the IEEE conference on computer vision and
  pattern recognition}, pages 770--778, 2016.

\bibitem{he2018amc}
Yihui He, Ji Lin, Zhijian Liu, Hanrui Wang, Li-Jia Li, and Song Han.
\newblock Amc: Automl for model compression and acceleration on mobile devices.
\newblock In {\em Proceedings of the European Conference on Computer Vision
  (ECCV)}, pages 784--800, 2018.

\bibitem{hluchyj1991shuffle}
Michael~G Hluchyj and Mark~J Karol.
\newblock Shuffle net: An application of generalized perfect shuffles to
  multihop lightwave networks.
\newblock {\em Journal of Lightwave Technology}, 9(10):1386--1397, 1991.

\bibitem{howard2019searching}
Andrew Howard, Mark Sandler, Grace Chu, Liang-Chieh Chen, Bo Chen, Mingxing
  Tan, Weijun Wang, Yukun Zhu, Ruoming Pang, Vijay Vasudevan, et~al.
\newblock Searching for mobilenetv3.
\newblock {\em arXiv preprint arXiv:1905.02244}, 2019.

\bibitem{howard2017mobilenets}
Andrew~G Howard, Menglong Zhu, Bo Chen, Dmitry Kalenichenko, Weijun Wang,
  Tobias Weyand, Marco Andreetto, and Hartwig Adam.
\newblock Mobilenets: Efficient convolutional neural networks for mobile vision
  applications.
\newblock {\em arXiv preprint arXiv:1704.04861}, 2017.

\bibitem{huang2017densely}
Gao Huang, Zhuang Liu, Laurens Van Der~Maaten, and Kilian~Q Weinberger.
\newblock Densely connected convolutional networks.
\newblock In {\em CVPR}, volume~1, page~3, 2017.

\bibitem{jia2014caffe}
Yangqing Jia, Evan Shelhamer, Jeff Donahue, Sergey Karayev, Jonathan Long, Ross
  Girshick, Sergio Guadarrama, and Trevor Darrell.
\newblock Caffe: Convolutional architecture for fast feature embedding.
\newblock In {\em Proceedings of the 22nd ACM international conference on
  Multimedia}, pages 675--678. ACM, 2014.

\bibitem{kandasamy2018neural}
Kirthevasan Kandasamy, Willie Neiswanger, Jeff Schneider, Barnabas Poczos, and
  Eric Xing.
\newblock Neural architecture search with bayesian optimisation and optimal
  transport.
\newblock {\em arXiv preprint arXiv:1802.07191}, 2018.

\bibitem{kawahara2009submodularity}
Yoshinobu Kawahara, Kiyohito Nagano, Koji Tsuda, and Jeff~A Bilmes.
\newblock Submodularity cuts and applications.
\newblock In {\em Advances in Neural Information Processing Systems}, pages
  916--924, 2009.

\bibitem{khosla2011novel}
Aditya Khosla, Nepali Jayadevaprakash, Bangpeng Yao, and Fei-Fei Li.
\newblock Novel dataset for fine-grained image categorization: Stanford dogs.

\bibitem{Kolmogorov:2002:EFM:645317.649315}
Vladimir Kolmogorov and Ramin Zabih.
\newblock What energy functions can be minimized via graph cuts?
\newblock In {\em Proceedings of the 7th European Conference on Computer
  Vision}, ECCV '02, pages 65--81, 2002.

\bibitem{krause2007near}
Andreas Krause and Carlos Guestrin.
\newblock Near-optimal observation selection using submodular functions.
\newblock In {\em AAAI}, volume~7, pages 1650--1654, 2007.

\bibitem{krizhevsky2009learning}
Alex Krizhevsky.
\newblock Learning multiple layers of features from tiny images.
\newblock Technical report, Citeseer, 2009.

\bibitem{li2017hyperband}
Lisha Li, Kevin Jamieson, Giulia DeSalvo, Afshin Rostamizadeh, and Ameet
  Talwalkar.
\newblock Hyperband: A novel bandit-based approach to hyperparameter
  optimization.
\newblock {\em The Journal of Machine Learning Research}, 18(1):6765--6816,
  2017.

\bibitem{liu2018progressive}
Chenxi Liu, Barret Zoph, Maxim Neumann, Jonathon Shlens, Wei Hua, Li-Jia Li, Li
  Fei-Fei, Alan Yuille, Jonathan Huang, and Kevin Murphy.
\newblock Progressive neural architecture search.
\newblock In {\em Proceedings of the European Conference on Computer Vision
  (ECCV)}, pages 19--34, 2018.

\bibitem{liu2017hierarchical}
Hanxiao Liu, Karen Simonyan, Oriol Vinyals, Chrisantha Fernando, and Koray
  Kavukcuoglu.
\newblock Hierarchical representations for efficient architecture search.
\newblock {\em arXiv preprint arXiv:1711.00436}, 2017.

\bibitem{liu2018darts}
Hanxiao Liu, Karen Simonyan, and Yiming Yang.
\newblock Darts: Differentiable architecture search.
\newblock {\em arXiv preprint arXiv:1806.09055}, 2018.

\bibitem{negrinho2017deeparchitect}
Renato Negrinho and Geoff Gordon.
\newblock Deeparchitect: Automatically designing and training deep
  architectures.
\newblock {\em arXiv preprint arXiv:1704.08792}, 2017.

\bibitem{paszke2017pytorch}
Adam Paszke, Sam Gross, Soumith Chintala, and Gregory Chanan.
\newblock Pytorch.
\newblock {\em Computer software. Vers. 0.3}, 1, 2017.

\bibitem{pham2018efficient}
Hieu Pham, Melody~Y Guan, Barret Zoph, Quoc~V Le, and Jeff Dean.
\newblock Efficient neural architecture search via parameter sharing.
\newblock {\em arXiv preprint arXiv:1802.03268}, 2018.

\bibitem{real2018regularized}
Esteban Real, Alok Aggarwal, Yanping Huang, and Quoc~V Le.
\newblock Regularized evolution for image classifier architecture search.
\newblock {\em arXiv preprint arXiv:1802.01548}, 2018.

\bibitem{sandler2018mobilenetv2}
Mark Sandler, Andrew Howard, Menglong Zhu, Andrey Zhmoginov, and Liang-Chieh
  Chen.
\newblock Mobilenetv2: Inverted residuals and linear bottlenecks.
\newblock In {\em Proceedings of the IEEE Conference on Computer Vision and
  Pattern Recognition}, pages 4510--4520, 2018.

\bibitem{shin*2018differentiable}
Richard Shin*, Charles Packer*, and Dawn Song.
\newblock Differentiable neural network architecture search, 2018.

\bibitem{szegedy2017inception}
Christian Szegedy, Sergey Ioffe, Vincent Vanhoucke, and Alexander~A Alemi.
\newblock Inception-v4, inception-resnet and the impact of residual connections
  on learning.
\newblock In {\em AAAI}, volume~4, page~12, 2017.

\bibitem{szegedy2015going}
Christian Szegedy, Wei Liu, Yangqing Jia, Pierre Sermanet, Scott Reed, Dragomir
  Anguelov, Dumitru Erhan, Vincent Vanhoucke, and Andrew Rabinovich.
\newblock Going deeper with convolutions.
\newblock In {\em Proceedings of the IEEE conference on computer vision and
  pattern recognition}, pages 1--9, 2015.

\bibitem{tan2018mnasnet}
Mingxing Tan, Bo Chen, Ruoming Pang, Vijay Vasudevan, and Quoc~V Le.
\newblock Mnasnet: Platform-aware neural architecture search for mobile.
\newblock {\em arXiv preprint arXiv:1807.11626}, 2018.

\bibitem{wang2019haq}
Kuan Wang, Zhijian Liu, Yujun Lin, Ji Lin, and Song Han.
\newblock Haq: Hardware-aware automated quantization with mixed precision.
\newblock In {\em Proceedings of the IEEE Conference on Computer Vision and
  Pattern Recognition}, pages 8612--8620, 2019.

\bibitem{wu2019fbnet}
Bichen Wu, Xiaoliang Dai, Peizhao Zhang, Yanghan Wang, Fei Sun, Yiming Wu,
  Yuandong Tian, Peter Vajda, Yangqing Jia, and Kurt Keutzer.
\newblock Fbnet: Hardware-aware efficient convnet design via differentiable
  neural architecture search.
\newblock In {\em Proceedings of the IEEE Conference on Computer Vision and
  Pattern Recognition}, pages 10734--10742, 2019.

\bibitem{xiong2019antnets}
Yunyang Xiong, Hyunwoo~J Kim, and Varsha Hedau.
\newblock Antnets: Mobile convolutional neural networks for resource efficient
  image classification.
\newblock {\em arXiv preprint arXiv:1904.03775}, 2019.

\bibitem{yang2018netadapt}
Tien-Ju Yang, Andrew Howard, Bo Chen, Xiao Zhang, Alec Go, Mark Sandler,
  Vivienne Sze, and Hartwig Adam.
\newblock Netadapt: Platform-aware neural network adaptation for mobile
  applications.
\newblock {\em Energy}, 41:46.

\bibitem{zhang2016understanding}
Chiyuan Zhang, Samy Bengio, Moritz Hardt, Benjamin Recht, and Oriol Vinyals.
\newblock Understanding deep learning requires rethinking generalization.
\newblock {\em arXiv preprint arXiv:1611.03530}, 2016.

\bibitem{Zhang_2018_CVPR}
Xiangyu Zhang, Xinyu Zhou, Mengxiao Lin, and Jian Sun.
\newblock Shufflenet: An extremely efficient convolutional neural network for
  mobile devices.
\newblock In {\em The IEEE Conference on Computer Vision and Pattern
  Recognition (CVPR)}, June 2018.

\bibitem{zoph2016neural}
Barret Zoph and Quoc~V Le.
\newblock Neural architecture search with reinforcement learning.
\newblock {\em arXiv preprint arXiv:1611.01578}, 2016.

\bibitem{zoph2017learning}
Barret Zoph, Vijay Vasudevan, Jonathon Shlens, and Quoc~V Le.
\newblock Learning transferable architectures for scalable image recognition.
\newblock {\em arXiv preprint arXiv:1707.07012}, 2(6), 2017.

\end{thebibliography}
